%% file: anonymous-submission-latex-2025.tex
\documentclass[letterpaper]{article} 
\usepackage{aaai25}  
\usepackage{times}  
\usepackage{helvet}  
\usepackage{courier}  
\usepackage[hyphens]{url}  
\usepackage{graphicx} 
\usepackage{natbib}  
\usepackage{caption} 
\usepackage{algorithm}
\usepackage{algorithmic}
\usepackage{booktabs}
\usepackage{xcolor}
\usepackage{multirow}
\usepackage{array}
\usepackage{tabularx}
\usepackage{tikz}
\definecolor{my1}{RGB}{142,207,201}
\definecolor{my2}{RGB}{255,190,122}
\definecolor{my3}{RGB}{250,127,111}
\definecolor{my4}{RGB}{130,176,210}
\definecolor{my5}{RGB}{35,186,197}
\usepackage{pgfplots}
\pgfplotsset{compat=1.17}
\usepackage{amssymb}
\usetikzlibrary{shapes.geometric}
\usetikzlibrary{arrows,decorations.pathmorphing,backgrounds,positioning,fit,petri}
\usetikzlibrary{arrows.meta,fit,shapes.arrows}
\usepackage{caption}
\usepackage{subfig}
\usetikzlibrary{positioning,shapes,arrows,shadows,patterns}
\usetikzlibrary{calc}
\usepackage{booktabs} 
\usepackage{multirow} 
\usepackage{arydshln}
\usepackage{pgfplots} 
\definecolor{darkgray}{RGB}{169,169,169}

\usepackage{tkz-kiviat,pgfplots}

\usepackage{enumitem}

\usepackage{newfloat}
\usepackage{listings}
\DeclareCaptionStyle{ruled}{labelfont=normalfont,labelsep=colon,strut=off} 
\floatstyle{ruled}
\newfloat{listing}{tb}{lst}{}
\floatname{listing}{Listing}
\title{NDP: \textbf{N}ext \textbf{D}istribution \textbf{P}rediction as a More Broad Target}

\author{
    Junhao Ruan\textsuperscript{\rm 1},
    Abudukeyumu Abudula\textsuperscript{\rm 1},
    Xinyu Liu\textsuperscript{\rm 1},
    Bei Li\textsuperscript{\rm 1}, 
    Yinqiao Li\textsuperscript{\rm 3}, \\
    Chenglong Wang\textsuperscript{\rm 1}, 
    Yuchun Fan\textsuperscript{\rm 1}, 
    Yuan Ge\textsuperscript{\rm 1},
    Tong Xiao\textsuperscript{\rm 1,2}\footnote{Corresponding author.} and
    Jingbo Zhu\textsuperscript{\rm 1,2}
}
\affiliations{
    \textsuperscript{\rm 1} School of Computer Science and Engineering, Northeastern University, Shenyang, China \\
    \textsuperscript{\rm 2} NiuTrans Research, Shenyang, China \\
    \textsuperscript{\rm 3} City University of Hong Kong \\
    rangehow@outlook.com,
    {\{xiaotong, zhujingbo\}@mail.neu.edu.cn}
}

\usepackage{bibentry}

\begin{document}
\nocopyright
\maketitle

\begin{abstract}
Large language models (LLMs) trained on next-token prediction (NTP) paradigm have demonstrated powerful capabilities. 
However, the existing NTP paradigm contains several limitations, particularly related to planned task complications and error propagation during inference.
In our work, we extend the critique of NTP, highlighting its limitation also due to training with a narrow objective: the prediction of a sub-optimal one-hot distribution. 
To support this critique, we conducted a pre-experiment treating the output distribution from powerful LLMs as efficient world data compression. 
By evaluating the similarity between the $n$-gram distribution and the one-hot distribution with LLMs, we observed that the $n$-gram distributions align more closely with the output distribution of LLMs.
Based on this insight, we introduce Next Distribution Prediction (NDP), which uses $n$-gram distributions to replace the one-hot targets, enhancing learning without extra online training time.
We conducted experiments across translation, general task, language transfer, and medical domain adaptation. 
Compared to NTP, NDP can achieve up to +2.97 COMET improvement in translation tasks, +0.61 average improvement in general tasks, and incredible +10.75 average improvement in the medical domain. This demonstrates the concrete benefits of addressing the target narrowing problem, pointing to a new direction for future work on improving NTP.

\end{abstract}

%

\section{Introduction}

Large Language Models (LLMs) are predominantly trained using the next-token prediction (NTP) paradigm. However, this approach has been subject to criticism, primarily focusing on two key issues: (1) the inability to perform tasks requiring advanced planning, such as look-ahead tasks \citep{kambhampati2024position,bachmann2024the}, and (2) error propagation during inference. These critiques have prompted various improvements, including methods to incorporate planning for future tokens during training or inference \citep{kambhampati2024llmscantplanhelp,monea2023passparallelspeculativesampling,chen2023acceleratinglargelanguagemodel,gloeckle2024better,cai2024medusasimplellminference}.

We posit that the NTP paradigm not only suffers from short-term thinking in the temporal dimension but also from a narrow candidate issue. During training, the model treats the successor token for a specific prefix as the sole correct target, attempting to approximate a one-hot distribution. This approach contrasts with human cognition, where multiple potential successor words are considered. Consequently, we argue that the probability distribution of successors should be non-one-hot.

Drawing inspiration from \citet{huh2024platonicrepresentationhypothesis}, who proposed that \textit{a model's ultimate representation should be a statistical model of underlying reality}, we consider that the ideal target for model learning is the statistical distribution over a comprehensive world data. We suggest that an $n$-gram statistical language model trained on such world data could provide this distribution efficiently. Furthermore, we hypothesize that even when trained on the same dataset as NTP use, $n$-gram LM can generate more efficient distributions than their NTP counterparts. To validate our hypothesis, we use LLM distributions as a proxy for the ideal statistical distribution of the world data, since LLM can be seen as a efficient compression of world data \citep{deletang2024language}. By comparing the similarities between $n$-gram distribution and one-hot distribution with LLM distribution on the same specific datasets, we demonstrate that $n$-gram distribution serves as a superior learning target since it aligns more with LLM distribution.

We introduce Next Distribution Prediction (NDP), an approach that draws inspiration from the $n$-gram LM concept to guide LLM training. This method calculates separate $n$-gram distributions for instruction and answer components, yielding supervised and causal language modeling (CLM) distributions, respectively. These distributions are then combined to replace the one-hot distribution during training.

Our extensive experiments across various models, tasks, and evaluation metrics demonstrate significant performance improvements. Moreover, NDP enables the simultaneous use of supervised and unsupervised data for training, effectively allowing for continued pre-training during fine-tuning. This feature is particularly advantageous for domain adaptation and language transfer scenarios. NDP outperforms NTP, showing improvements of up to 2.97 COMET points in translation tasks, an average gain of 0.61 points in general tasks, and a remarkable average increase of 10.75 points in the medical domain.

Our primary contributions are:

\begin{itemize}
\item A novel critique of NTP, highlighting its limitations in both temporal and candidate space dimensions, providing a foundation for future improvements.
\item The introduction of NDP, a preliminary solution that achieves substantial performance gains across a wide range of tasks and model configurations without additional online training overhead. To our knowledge, this is the first method to explore $n$-gram LM for supervised LLM training.
\item An innovative approach integrating unsupervised data into supervised training with NDP, aiming to combine continued pre-training and instruction fine-tuning, thus enhancing LLM adaptability to specific domains and languages.
\end{itemize}

\section{Related Work}

\subsection{Knowledge Distillation}
Both Knowledge Distillation (KD) and NDP employ non-one-hot distributions as targets. These two approaches, however, can be considered as parallel dimensions of work. Fundamentally, the teacher models employed in neural network-based knowledge distillation methods must still be obtained through statistical NTP or NDP training paradigms. This inherent dependency means that neural network-based knowledge distillation cannot supersede NTP or NDP; rather, they are complementary. In practical applications, the typical LLM training process involves NTP-based pre-training and instruction fine-tuning, optionally followed by teacher network-based knowledge distillation. These are distinct processes that can be applied sequentially to enhance the student network's quality. From a training cost perspective, teacher network-based knowledge distillation introduces a teacher model with typically more than 10 times the parameters of the student model, significantly increasing both training time and hardware requirements.

NDP and NTP can be conceptualized as forms of dataset distillation, differing in their level of granularity: token-level for NDP and sentence-level for NTP. This perspective illuminates NTP's limitations. Yuan et al. (2023) demonstrated that in knowledge distillation, student models more readily assimilate soft labels compared to one-hot labels. Wei et al. (2024) observed that the efficacy of sentence-level versus token-level distillation correlates with student model size, with larger models benefiting more from token-level approaches. Empirically, most research utilizing black-box Large Language Models (LLMs), such as instruction data synthesis (Xu et al., 2023), employs sentence-level distillation. While effective, sentence-level distillation alone has not enabled open-source LLMs to match the performance of GPT-4-turbo/GPT-4 (OpenAI, 2024). Conversely, Gemma2-9B (Gemma Team, 2024) achieved performance comparable to LLaMA3-8B (Dubey et al., 2024) with only 9T pretraining tokens, attributable to its use of token-level distillation. These findings support NDP's superior performance over NTP.

\subsection{Calibration During Training}
Our work shares similarities with output probability calibration methods, as both aim to mitigate overconfidence and align output probabilities with true probabilities. Prominent calibration techniques during training include loss function modification (Ren et al., 2024; Li et al., 2020; Lin et al., 2018) and label smoothing (Liang et al., 2024; Wei et al., 2022; Malagutti et al., 2024).

Research on loss function modification often attributes the discrepancy between predicted and real-world probabilities to maximum likelihood estimation. This has led to efforts to replace cross-entropy (based on negative log-likelihood) with alternative loss functions, introducing significant computational overhead and sensitive parameters. In contrast, NDP can be easily integrated into existing training frameworks without incurring additional training costs, yielding substantial improvements.

While NDP supports smoothing, its primary advantage stems from addressing the issue of narrow candidates rather than smoothing per se. NDP guarantees a non-one-hot distribution, allowing for multi-discrete value distributions rather than only continuous ones. Given the expanding vocabulary sizes in modern language models, the correct next token candidates cannot span the entire vocabulary range. For large language models requiring high-precision alignment, introducing noise across the entire vocabulary can result in downstream task performance inferior to that achieved with one-hot distributions from NTP.

\subsection{Improvement on Next Token Prediction}

Earlier criticisms of the NTP training paradigm were all focused on the time dimension, which led to many improvements.
\citet{monea2023passparallelspeculativesampling} was inspired by Speculative Sampling \citep{chen2023acceleratinglargelanguagemodel}, using the LLMs itself as a draft model, thus allowing the LLMs to output multiple tokens at once during the inference stage, implicitly achieving long-term planning and alleviating the short-term issues to some extent. \citet{gloeckle2024better} achieved consistent improvements in efficiency and performance on code tasks by training shared model backbones and multiple independent output heads and adopting speculative decoding with Medusa-like tree attention \citep{cai2024medusasimplellminference} during inference, indicating that this training paradigm has advantages in large-scale models.

These works are completely orthogonal to our perspective. We primarily focus on the issues brought by narrow candidates, with the hope of jointly optimizing the NTP process.

\section{Preliminary Experiments}

\begin{figure}
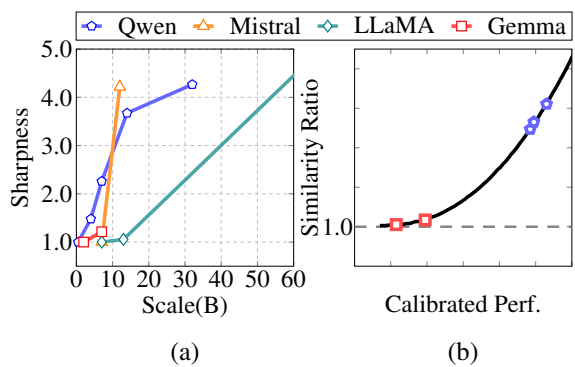

    \centering
    \include{figures/sharpandsimilarity}

    \caption{(a): Changes in the sharpness of model distributions with increasing model size.
    (b): Changes in $\texttt{Sim}_{ngram}$/$\texttt{Sim}_{ntp}$ with calibrated preformance increases.}
    \label{fig:sharpandsimilarity}
\end{figure}

We aim to define the strengths of learning targets by exploring the similarity between targets and world data distributions. We use the distribution of LLM to approximate the representation of real-world data distributions, as LLM can be seen as an efficient compression of world data \citep{deletang2024language}. The learning targets we explore are primarily the NTP distribution and the $n$-gram distribution, which is derived by statistically obtaining global information on specific datasets.

We explore the cosine similarity $\texttt{Sim}_{ngram}$ between the distributions generated by $n$-gram LM and LLM, as well as $\texttt{Sim}_{ntp}$ generated by NTP and LLM. We conduct experiments on various LLMs and datasets, which are listed in Appendix A.1.
We mainly want to observe how $\texttt{Sim}_{ngram}$ and $\texttt{Sim}_{ntp}$ change as the LLM performance improves (since we believe LLM with better performance compress world data more effectively). We observe two interesting phenomena.
\begin{itemize}
    \item The sharpness\footnote{We employ two metrics: one is the proportion of elements required for the distribution to reach a probability $p$. For a very sparse one-hot distribution, its proportion will always be $1/|V|$, where V is the vocabulary of the model. The other is kurtosis, which we demonstrate in Appendix A.1.} of the model distribution increases rapidly with the scale of the model, as shown in Figure 1(a).
    \item No matter which LLM is used as world data proxy, we consistently observe that 
$\texttt{Sim}_{ngram}$ is greater than $\texttt{Sim}_{ntp}$. Moreover, as the LLM's performance\footnote{After calibration, We choose the average score list on hugging face Open LLM leaderboard as metrics of performance.} improves, the ratio $\texttt{Sim}_{ngram}$/$\texttt{Sim}_{ntp}$ approaches exponential growth.
\end{itemize}

Notably, sharpness introduces a bias to the observation of extremely sharp one-hot NTP distributions. To eliminate this interference, we divide the model's performance by its sharpness as a calibration for performance. 
To unify the observation of various LLMs, we use ratio $\texttt{Sim}_{ngram}$/$\texttt{Sim}_{ntp}$ rather than absolute value as the vertical axis  in Figure \ref{fig:sharpandsimilarity}(b).

The series of results sufficiently demonstrates that the $n$-gram distribution better approximates the distribution of world data, therefore making it a more promising learning target compared to the NTP distribution.

We also conducted a regression analysis on sharpness, ratio, and performance, consistently finding a positive correlation between ratio and performance. This further supports the idea that $n$-gram distribution better aligns with world data distribution. More details are provided in Appendix A.1.

\section{Next Distribution Prediction Paradigm}

\begin{figure*}[t]
\centering
\includegraphics[width=0.9\textwidth]{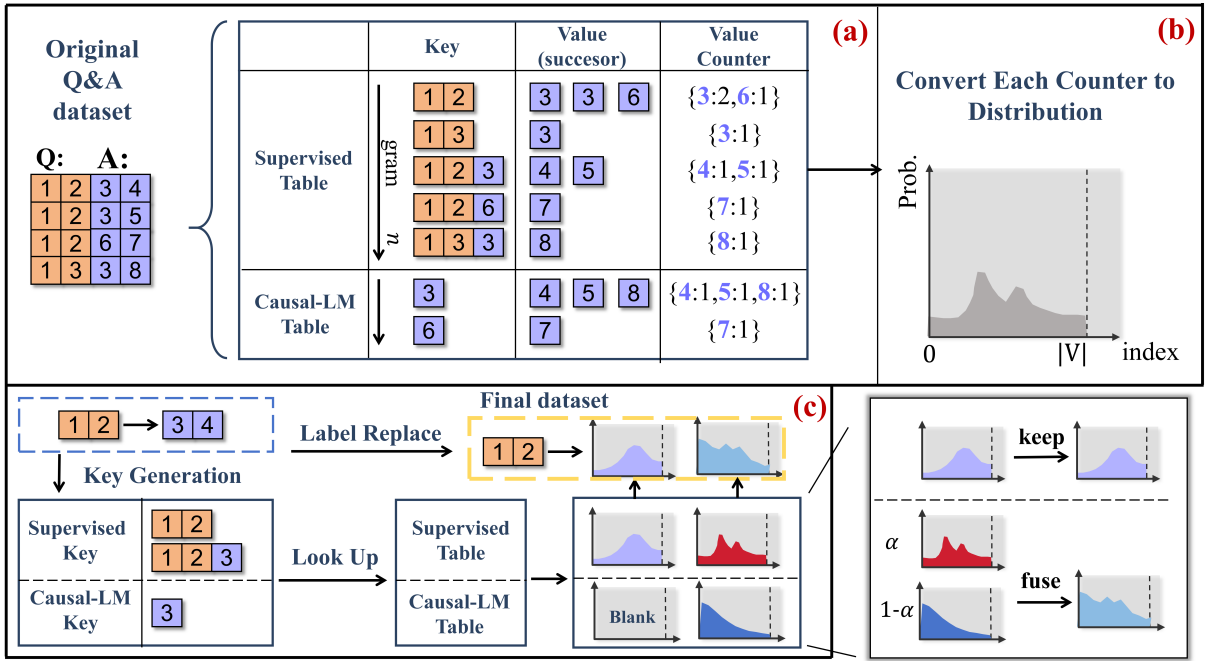} 
\caption{Overall framework of NDP. The numbers in the squares represent the token index in vocabulary. The $n$-grams required for the supervised table are counted starting from the question/input, meaning the counting begins from \textcolor[cmyk]{0.0519,0.3803,0.5142,0.00}{\rule{1.5ex}{1.5ex}}. The CLM table only counts from the answer/output, which means the counting starts from \textcolor[cmyk]{0.27,0.36,0,0}{\rule{1.5ex}{1.5ex}}.}
\label{fig:framework}
\end{figure*}

 In this section, we will provide a detailed description of how our method, NDP, incorporates the aforementioned statistical $n$-gram concept into the actual model training process. 
 
Almost all datasets can be categorized as either unsupervised or supervised datasets. Let's take supervised datasets as an example, since self-supervised datasets can be regarded as a special case of supervised datasets where the instruction/input is empty. Based on the training forms of pre-training and instruction fine-tuning, we aim to extract two distributions from the dataset: the Causal Language Modeling (CLM) distribution and the supervised distribution. 

This process can be divided into three sub-processes: First, learn the $n$-gram table through statistical analysis of the dataset (Figure \ref{fig:framework}(a)). Second, convert value counter from the $n$-gram table to distribution (Figure \ref{fig:framework}(b)). And third, replace the training targets in the original dataset from one-hot distributions to non-one-hot distributions (Figure \ref{fig:framework}(c)). 

\subsection{Learning $n$-gram Tables}


 The specific process is illustrated in Figure \ref{fig:framework}(a). Given a sentence, we use all its $n$-grams as keys and the corresponding successor tokens as values to form several key-value pairs. Across the entire training set, the key-value pairs formed by different sentences are merged based on the identical key, and corresponding values will collectively form a frequency Counter. The statistical processes for the supervised table and the CLM table are independent. The supervised table gathers supervised information and starts counting $n$-gram from the question part, while the CLM table requires only unsupervised information, so it counts only from the answer part.

It is easy to observe that the one-hot distribution derived from NTP is actually a special form of supervised table. When none of the keys overlap in the table generated from the entire dataset, the distribution derived from the supervised table will be completely equivalent to NTP distribution. At the same time, considering that the key length in CLM is always shorter than in supervised, the corresponding values will be significantly more frequent after mixing different sentences compared to the supervised distribution. To mitigate this potential effect, where overly abundant language modeling information might dilute the information from supervised data, we empirically begin to compile the CLM table from 5-gram based on some simple experimental results. When performing $n$-gram statistics solely on an unsupervised dataset, we only compile the CLM table and start from 1-gram.

\subsection{Converting Distributions from $n$-gram Tables}
In this subsection, we will introduce how to complete Figure \ref{fig:framework}(b).
Figure \ref{fig:framework}(b) shows the process of converting each element in the value counter into a distribution on the model vocabulary dimension. For each counter, We create a tensor with the dimensions of the vocabulary, extracting the indices and corresponding counts from the frequency counter and setting the tensor accordingly. Then, we convert this tensor into a probability distribution using softmax.

It is important to note that the vocabulary's dimension is typically much larger than the number of items in the Counter, for instance, 256k vs. 10. If we directly form a probability distribution on such a frequency vector, it would result in a uniform distribution, diluting the information derived from the dataset. We solve this problem by controlling the probability allocated to the zero regions of the tensor. 

Without loss of generality, we rearrange a vocab tensor $v=[a_1,a_2...a_{|V|}]$ with $|V|$ elements into two contiguous regions based on whether the elements are zero or non-zero. This rearrangement results in $v'=[a'_1...a'_k,a'_{k+1}...a'_{|V|}]$, where $[a'_1...a'_k]$ represents the non-zero elements region, and $[a'_{k+1}...a'_{|V|}]$ represents the zero elements region. Therefore, the softmax process on $v'$ can be described as Equation \ref{equ1}.

\begin{equation}
\label{equ1}
\texttt{Softmax}(v')=\frac{\sum_{i=1}^k{e^{a'_i}}}{\sum_{j=1}^{|V|}{e^{a'_j}}}+\frac{\sum_{i=k+1}^{|V|}{e^{a'_i}}}{\sum_{j=1}^{|V|}{e^{a'_j}}}
\end{equation}

\noindent where the second item in Equation \ref{equ1} is the probability value allocated to the entire zero elements region. To control its value and make it equal to our preset probability 
$p$, we introduce a temperature coefficient 
$t$, transforming it to solve Equation \ref{equ2}.
\begin{equation}
\label{equ2}
\texttt{f}(t)=\frac{\sum_{i=k+1}^{|V|}{e^{a'_i/t}}}{\sum_{j=1}^{|V|}{e^{a'_j/t}}}=p
\end{equation}
Obtaining an exact solution for Equation 2 is quite challenging; however, we can easily obtain its approximate solution through numerical computation methods. For instance, the root-finding methods provided in scipy\footnote{https://docs.scipy.org/doc/scipy/tutorial/optimize.html\#root-finding}, or the simpler bisection method, can efficiently locate $t$ within the [0, 100] interval, with the error easily controlled within 1e-6.

\subsection{Replacing Origin One-hot Target}
After properly handling the token-level distribution, we can simply traverse the original dataset to replace the training targets. In Figure \ref{fig:framework}(c), we provide an example with a sentence.
First, we decompose a sentence into corresponding keys as in Figure \ref{fig:framework}(a). Then, we use the keys to look up the corresponding table and obtain the distribution that we transformed in Figure \ref{fig:framework}(b). Because the CLM table starts collecting statistics only after the first token, the first token itself can only yield a blank distribution. We

For the supervised and CLM distributions of the same token, we employ a simple linear weighted fusion. Fusion allows us to combine the strengths of both distributions and avoids the overfitting and extra training costs in separate training phases. We illustrate this process in Equation \ref{equ3}.

\begin{equation}
\label{equ3}
D_{mix}=\alpha D_{supervised}+(1-\alpha)D_{CLM}
\end{equation}

where $\alpha$ is a hyperparameter constrained to the interval [0, 1]. We substitute the original one-hot label target with $D_{mix}$. It is important to note that when encountering a blank distribution in the fusion objects, we do not perform fusion but instead retain the original distribution. In other words, we consistently assign zero weight to blank distributions during the fusion process. 

At this point, we have completed the data processing part. The next step in the training process is the regular teacher-forcing as NTP.
NDP has another very interesting use: we can use a large amount of unlabeled text to further enhance the CLM table. This process essentially unifies pre-training and fine-tuning. We will demonstrate this in experiments later.


\section{Experiments}
\subsection{General Tasks for Large Language Models}
In this subsection, we aim to explore the impact of using NDP for instruction fine-tuning (IFT) on the base model for general tasks. The specific experimental setup is as follows.
\setlength{\tabcolsep}{1.5mm}
\begin{table*}[htbp]
\aboverulesep=-1pt
 \centering
\begin{tabular}{l | c c c c c c c c c c c | c c}
\toprule
& GSM & MMLU & HE &  TruQA & BBH & ARC-C & TriQA  &AE & SCIQ & WG & IFeval & Avg. \\
\midrule 
Gemma& 19.56 & 42.12 & 23.78 & 33.05 & 35.94 & 40.36 & 47.38 & 27.20 & 94.30 & 65.67 & 14.42 & 40.34 \\
\cdashline{2-13}
 \addlinespace[2pt]
\hspace{2.5mm}+NTP & \textbf{24.49} & 40.85 & 31.71 & 41.59 & 37.40 & 44.54 & \textbf{42.73} & 27.54 & 92.30 & \textbf{66.61} & 19.41 & 42.65 \\
\hspace{2.5mm}+LS & 22.37&40.90&30.49&40.29 &35.80 	&44.11 	&40.52 	&27.27 	&\textbf{93.30} &66.46 	&13.12 	&41.33 \\
\hspace{2.5mm}+NDP & 23.81 & \textbf{41.34} & \textbf{35.98} & \textbf{42.84} & \textbf{37.77} & \textbf{44.88} & 42.57 & \textbf{28.65} & 91.90 & 66.38 & \textbf{19.78} & \textbf{43.26} \\
\midrule
\midrule
LLaMA3& 54.44 & 65.57 & 37.20 & 43.91 & 62.52 & 50.43 & 71.21 & 33.70 & 96.30 & 78.22 & 10.17 & 54.88 \\
\cdashline{2-13}
 \addlinespace[2pt]
\hspace{2.5mm}+NTP & \textbf{53.30} & 62.37 & 37.80 & 44.04 & 60.13 & \textbf{51.28} & \textbf{63.64} & 33.38 & \textbf{96.60} & 77.82 & 14.97 & 54.12 \\
\hspace{2.5mm}+LS & 50.80 & 58.23 & 35.37 & 43.73 & 51.14 & 51.02 & 57.69 & \textbf{33.51} & 96.50 & \textbf{78.22} & 17.19 & 52.13 \\
\hspace{2.5mm}+NDP & 53.15 & \textbf{62.99} & \textbf{39.02} & \textbf{44.05} & \textbf{61.86} & 51.11 & 62.70 & 33.24 & 96.50 & 77.98 & \textbf{17.93} & \textbf{54.59} \\

\bottomrule
\end{tabular}
\caption{Evaluation results on general tasks. The benchmark abbreviations in the table: GSM (GSM8k), MMLU (Massive Multitask Language Understanding), HE (HumanEval), TruQA (TruthfulQA), BBH (Big-Bench Hard), ARC-C (ARC-Challenge), TriQA (TriviaQA), AE (AgiEval), SCIQ (SCIQ), WG (WinoGrande), and IFeval (IFeval). LS means Label Smoothing here.}
\label{tab:performance}
\end{table*}

\subsubsection{Model \& Baseline}
We conducted experiments on Gemma-2B \citep{gemmateam2024gemmaopenmodelsbased} and LLaMA3-8B \citep{dubey2024llama3herdmodels}. We used LoRA \citep{hu2022lora} to train LLaMA3-8B. We mainly compare NDP with NTP, label smoothing and knowledge distillation (KD). In KD we choose Gemma2-27B as the teacher model to teach Gemma-2B. The comparative experiments between NDP and knowledge distillation(KD) are listed in Appendix A.2 .

\subsubsection{Dataset}
We selected a mixture of Alpaca-GPT4 \citep{peng2023instruction}, Math \citep{hendrycksmath2021}, and Code \citep{zheng2024opencodeinterpreterintegratingcodegeneration} as the instruction fine-tuning (IFT) dataset. This combination is similar to the typical mix of general text, code, and math used in pretraining, with a total dataset size of 220K instances. 
The evaluation comprised 11 benchmarks that broadly cover the model's general reasoning, knowledge Q\&A, mathematics, coding, fact, and instruction following capabilities. More detailed benchmark information can be found in Appendix A.2.

\subsubsection{Evaluation framework}
 Our evaluation process primarily leveraged the lm-evaluation-harness framework \citep{eval-harness}, with the exception of coding tasks, for which we utilized the evaluation scripts from the OpenAI/HumanEval repository. 
The evaluation setting closely follow those outlined in the LLaMA3 evaluation protocol\footnote{https://github.com/meta-llama/llama3/blob/main/eval\_details}.

\subsubsection{Results}
The experimental results are summarized in Table \ref{tab:performance}. From the result, we observe some phenomena:
\begin{itemize}
    \item \textbf{Consistent Improvements of NDP}. NDP outperforms NTP by +0.61 points on Gemma and by +0.47 points on LLaMA, respectively. These improvements validate the efficacy of the modifications introduced to address the inherent limitations of the NTP paradigm.
    
    \item \textbf{Failure of Label Smoothing}. Contrary to expectations, the label smoothing technique did not yield performance gains. In fact, it underperforms relative to the NTP method. We hypothesize that the meaningless noise during the instruction fine-tuning phase may have degraded the quality of the fine-tuning data. This observation indicates that the critical importance of data quality over quantity in the instruction fine-tuning process.
    \item \textbf{Drop of LLaMA3 Models.} NTP and NDP Both show a slight decline compared to the LLaMA3-base model. We believe the possible reasons for this are: the pre-training data of LLaMA3-8b amounts to an astonishing 15T tokens, and despite some deduplication, there is still a significant possibility of data leakage in the benchmark. The same phenomenon also appeared in the work of \cite{xu2024magpie}, which listed results where fine-tuning with various mainstream instruction data caused a decline in benchmark performance.  
\end{itemize}
\subsection{Translation Task for Encoder-Decoder Models}

\begin{table*}[ht]
    \centering 
    \begin{tabular}{llcccccccc} 
        \toprule 
        & & \multicolumn{2}{c}{IWSLT17} & \multicolumn{2}{c}{WMT22} & \multicolumn{2}{c}{Avg.} & \multicolumn{2}{c}{Avg. $\Delta$} \\
        \cmidrule(r){3-4} \cmidrule(r){5-6} \cmidrule(r){7-8} \cmidrule(r){9-10} 
        & & BLEU & COMET & BLEU & COMET & BLEU & COMET & BLEU & COMET \\
        \midrule 
        \multirow{2}{*}{T5\_small} 
        & NTP & 11.51 & \textbf{55.49}& 7.39 & \textbf{48.43}& 9.45 & \textbf{51.96} & \multirow{2}{*}{\textbf{+0.30}} & \multirow{2}{*}{-0.18} \\
        & NDP & \textbf{11.56}& 55.39 & \textbf{7.93}& 48.16 & \textbf{9.75} & 51.78 &  &  \\
        \midrule 
        \multirow{2}{*}{T5\_base} 
        & NTP & 19.97 & 68.54 & \textbf{15.48}& 61.90 & 17.73 & 65.22 & \multirow{2}{*}{\textbf{+0.91}} & \multirow{2}{*}{\textbf{+1.97}} \\
        & NDP & \textbf{21.87}& \textbf{70.66}& 15.41 & \textbf{63.72}& \textbf{18.64} & \textbf{67.19} &  &  \\
        \midrule 
        \multirow{2}{*}{T5\_large} 
        & NTP & 23.63 & 76.42 & 17.49 & 71.26 & 20.56 & 73.84 & \multirow{2}{*}{\textbf{+1.96}} & \multirow{2}{*}{\textbf{+1.17}} \\
        & NDP & \textbf{25.70}& \textbf{77.29}& \textbf{19.34}& \textbf{72.72}& \textbf{22.52} & \textbf{75.01} &  &  \\
        \bottomrule 
    \end{tabular}
    \caption{T5 series evaluated on IWSLT17 \& WMT22 with BLEU and COMET22.} 
    \label{tab:t5} 
\end{table*}

\begin{table*}[htbp]
\aboverulesep=-1pt 
 \centering
\begin{tabular}{l | c c c c | c c | c| c}
\toprule
& MedQA & MedMCQA & PubMedQA & CareQA  & Avg.  & Avg. $\Delta$ & MMLU & Flos\\ 
\midrule 
Qwen2&44.46 &46.57 &47.30 &52.04 &47.59 & - & 70.76 & -\\
\cdashline{2-9}
 \addlinespace[2pt]
\hspace{2.5mm}+CPT$^{\ddagger}$+NTP &46.58  &45.76 &24.30 &56.48 &43.28  & -4.31 & 68.18 & $6.44\times10^{19}$ \\
\hspace{2.5mm}+NTP &47.60 &50.11 &42.60 &60.47 &50.19  & +2.60 & 70.97& $1.72\times10^{18}$\\
\hspace{2.5mm}+NDP &49.49 &50.68&42.10&59.95 &50.83 & +3.24  & \textbf{71.00} &$1.71\times 10^{18}$\\
\hspace{2.5mm}+NDP$^\dagger$ &\textbf{49.49} &\textbf{50.83} &\textbf{43.70}  &\textbf{61.22}&\textbf{51.25} & \textbf{+3.66} & 70.99 &$1.71\times 10^{18}$\\
\midrule 
\midrule 
LLaMA3&33.70 &36.22 &2.50 &46.98 &29.85 & - & 65.57 & -\\
\cdashline{2-9}
 \addlinespace[2pt]
 \hspace{2.5mm}+CPT$^{\ddagger}$+NTP &25.29 &37.25 &10.00 &48.76 &30.33  & +0.48 & 58.43&  $6.62\times10^{19}$ \\
\hspace{2.5mm}+NTP &31.26 &33.09 &13.40 &39.25 &29.25  & -0.60 & 54.09& $1.75\times10^{18}$\\
\hspace{2.5mm}+NDP &20.27 &27.40&\textbf{53.7}&22.99 &31.09 & +1.24  & 54.77 &$1.74\times 10^{18}$\\
\hspace{2.5mm}+NDP$^\dagger$ &\textbf{38.41} &\textbf{39.61} &31.6  &\textbf{50.36}&\textbf{40.00} & \textbf{+10.15} & \textbf{58.58} &$1.74\times 10^{18}$\\
\bottomrule
\end{tabular}
\caption{ Results on domain adaptation task. Item marked with $\dagger$ represents enhancement using PubMed, while $\ddagger$ means Pubmed+Redpajama. }
\label{tab:medical}
\end{table*}

In this subsection, we aim to answer the following questions: (1) Does our method work effectively for models with smaller parameter sizes?  (2) Can our method benefit specific downstream tasks? Although we have demonstrated that NDP can benefit general, broad tasks, further discussion on adaptation to specific task can still be argued. 

\subsubsection{Model \& Baseline} The T5 model \citep{raffel2023exploringlimitstransferlearning} is an excellent choice because we will select a 400M decoder-only LLaMA in the latter experiment. Using T5 would allow us to observe the impact on encoder-decoder models as well. We selected three sizes of the T5 1.1 version models: small (77M), base (248M), and large (783M). Here we only compare with NTP, since in preliminary experiments, label smoothing has already shown a similar drop in performance as general tasks.
\subsubsection{Dataset \& Metric}
 We selected 200k bilingual sentence pairs in the en-de direction from IWSLT17 \citep{cettolo-etal-2017-overview} as the training set. Both IWSLT17 and WMT22 \citep{kocmi-etal-2022-findings} were used as test sets, as IWSLT17 consists of TED talk utterance transcripts while WMT22 comprises news articles. We used WMT22 to observe the generalization performance on out-of-domain data. We use rule-based 
 SacreBLEU \citep{post-2018-call} and neural network based COMET22 \citep{rei-etal-2022-comet} as evaluation metrics. 

\subsubsection{Results}
Translation result is shown in Table \ref{tab:t5}. Overall, NDP consistently outperformed NTP in both in-domain and out-of-domain performance except COMET on T5-small. This suggests that NDP also has considerable potential in small models and downstream-specific tasks.

\subsection{Unifying Continue Pre-training and Fine-tuning}

\begin{figure}[t]
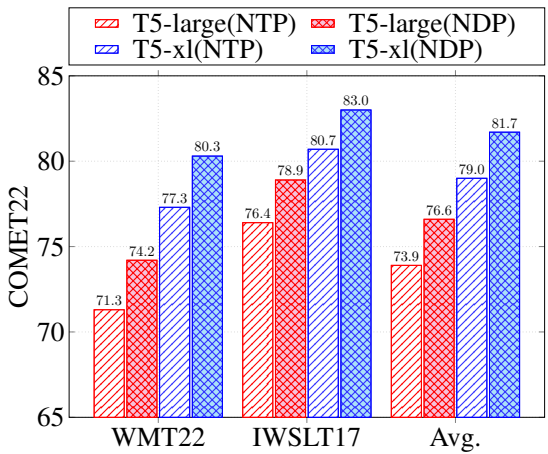

  \centering
    \include{figures/domainpicture}

    \caption{Comparison of COMET22 scores for different models on WMT22 and IWSLT2017 datasets} 
    \label{fig:mono-translate} 
\end{figure}

Post-training of large language models often includes continued pertaining (CPT), IFT, and RLHF. Here, we focus on CPT and IFT since they involve NTP. Post-training is usually a delicate and complex process because the goal is not only to adapt to a specific domain or align with humans but also to ensure that the knowledge learned during pre-training is minimally disrupted. The NDP offers an optional approach. 
We still use the dataset from the IFT phase to generate the $n$-gram table, but we additionally use the dataset from the CPT phase to enrich the CLM table, making the resulting CLM distribution incorporates information from the CPT tasks and become more robust.
Our method has the following three potential benefits in unifying CPT and IFT: 1) it avoids the cumbersome hyperparameter selection during the continued pre-training phase, such as learning rate, decay, and warmup. 2) by forgoing the continued pre-training phase, the number of model update steps is greatly reduced, which helps alleviate the problem of model forgetting. 3) it saves a lot of training resources since the model still computes only the original instruction dataset.

We have selected two common scenarios: \textbf{language transfer} and \textbf{domain adaptation}. 

\subsubsection{Language Transfer} We extracted 500k sentences from the monolingual German data in WMT23 \citep{kocmi-etal-2023-findings} to supplement the CLM distribution. We use T5-1.1-3B and T5-1.1-Large as the models for this experiment. The COMET result is shown in Figure \ref{fig:mono-translate}, while BLEU result in Appendix A.3.

Our approach achieved a gain of +2.72 on T5-large model and +2.64 on the t5-XL model compared to NTP on average. The BLEU score increased even more, with a gain of +3.18 on the T5-large model and +2.68 on the T5-XL model, confirming the significant benefits of NDP in unifying CPT and IFT. 
For comparison, similar attempts have been made by NLLB \citep{nllbteam2022languageleftbehindscaling}, which employed an encoder-decoder model with a Denoising Autoencoder (DAE) as the pre-training task. This task, akin to a cloze test, is simpler than CLM. Additionally, NLLB performs unifying at a step-wise granularity, alternating between fully computing the DAE and the fine-tuning loss in separate steps. In contrast, ours integrates the losses from both the CLM task and the supervised task at a loss-wise level within each iteration, providing a more fine-grained approach.

\subsubsection{Domain Adaptation}
We choose PubMed\_Abstract which sampled from pile \citep{gao2020pile800gbdatasetdiverse} and Redpajama-1B as CPT dataset and Alpaca\_GPT4+Medquad \citep{Ben_Abacha_2019} as IFT dataset. Test on following benchmarks: MedQA \citep{jin2021disease}, MedMCQA \citep{pmlr-v174-pal22a}, CareQA \citep{gururajan2024aloe}, MMLU \citep{hendryckstest2021}, and PubMedQA \citep{jin2019pubmedqa}. We retain  MMLU to observe its impact on the general domain indirectly. We do full parameter tuning on Qwen2-7B \citep{qwen} and LLaMA3-8B. The result is shown in Table \ref{tab:medical}, and we can observe that Qwen2 has undergone more adaptation in the medical field compared to LLaMA3. The key findings are as follows:
\begin{itemize}
    \item For models that lack domain-specific pre-training (such as LLaMA3), NTP leads to a performance drop. In contrast, NDP maintains a steady performance increase.
    \item The advantages of domain data augmentation are more evident in models without extra domain-specific pre-training. Specifically, Qwen2 exhibits an improvement of +3.66, while LLaMA3 shows a significant increase of +10.15. This suggests that our method holds substantial potential for enhancing the unified continued pretraining process. Moreover, CPT benefits LLaMA3 but negatively impacts Qwen2.
    \item Across all settings observed in the MMLU benchmark, models trained with NDP not only show superior domain adaptation but also match the general capabilities of models trained with NTP.
\end{itemize}
More training details can be found in Appendix A.4.



\subsection{The Convergence Endpoints of NDP and NTP}

\begin{figure}[t]
\centering
\includegraphics[width=0.9\columnwidth]{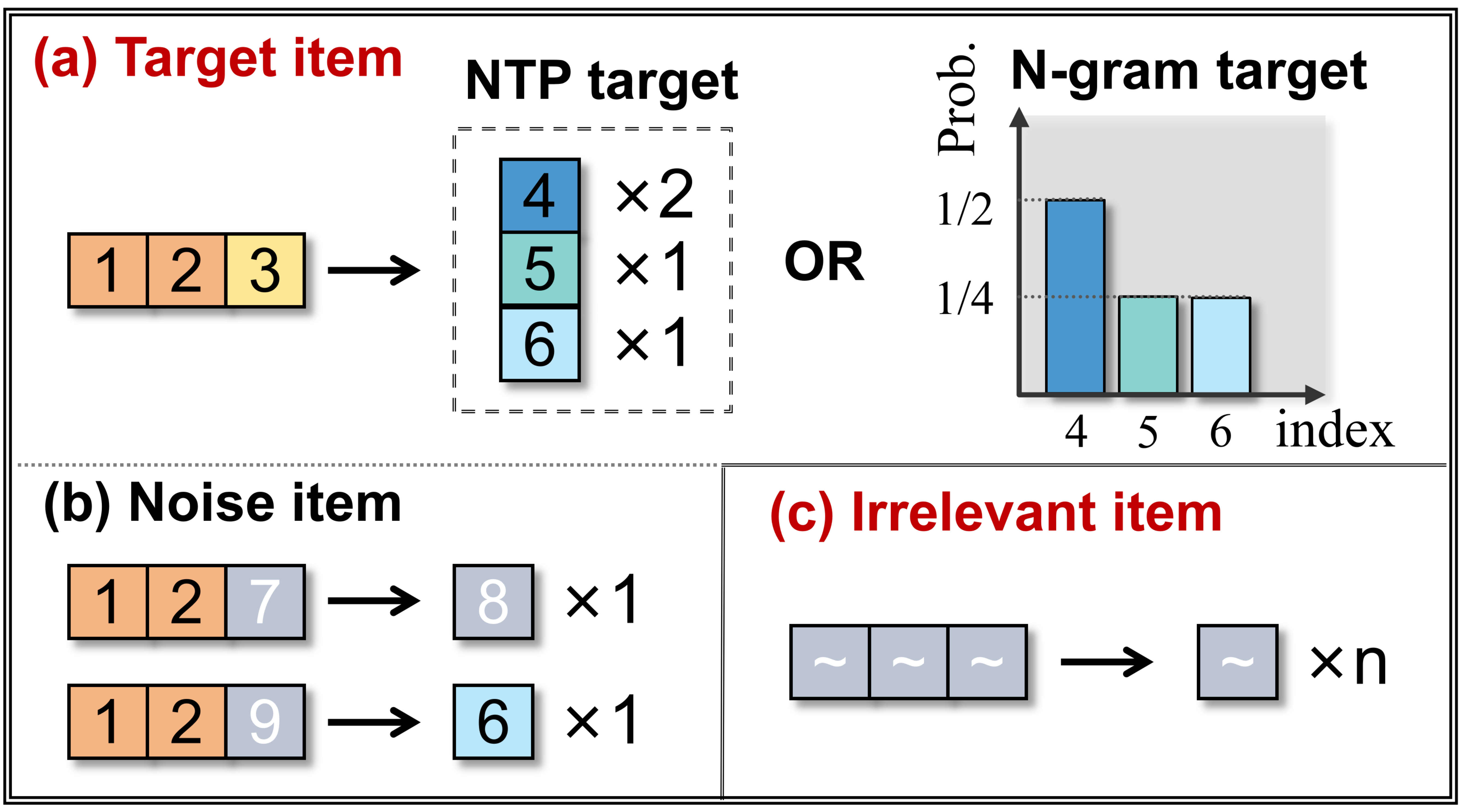} 
\caption{ Dataset configuration. Numbers represent tokens, and tilde represents a token that does not repeat with other tokens. The items marked in red font indicate that we will observe its fitting accuracy. \textcolor[RGB]{245,180,130}{\rule{1.5ex}{1.5ex}} are used to represent the common prefix of the input, \textcolor[RGB]{254,230,149}{\rule{1.5ex}{1.5ex}} represent the different suffixes of the input, the blue blocks represent the target to be predicted, and \textcolor[RGB]{189,196,212}{\rule{1.5ex}{1.5ex}} represent the irrelevant tokens. $n=40$ in our setting.}
\label{fig:datasetconfig}
\end{figure}

\begin{figure}[t]
\centering
\includegraphics[width=0.45\textwidth]{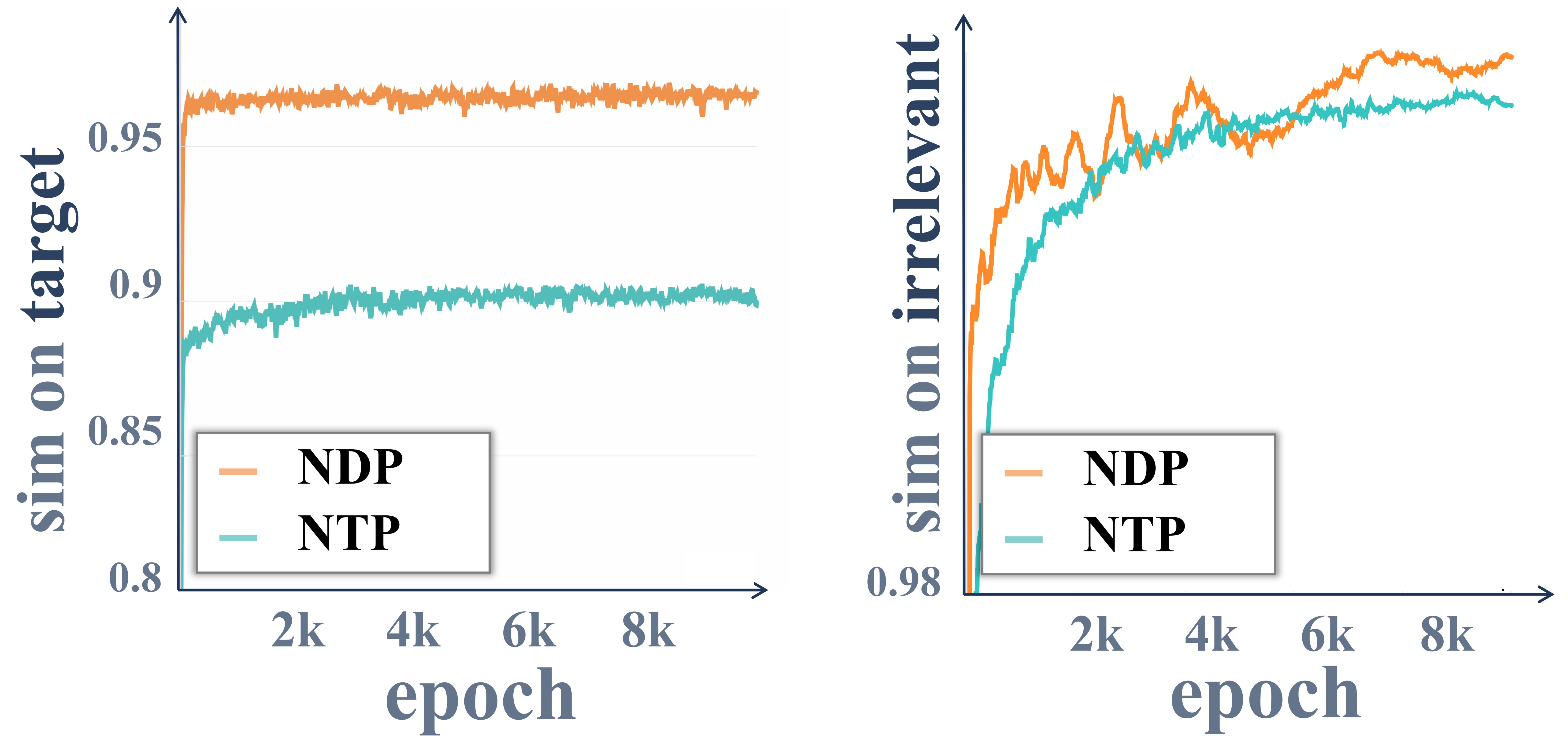} 
\caption{Analysis of model convergence with increasing training epochs. The left figure shows the similarity of the model's output distribution on the target items. The right figure shows the similarity with irrelevant items.}
\label{fig:similarity_distribution}
\end{figure}
NDP demonstrates a notable advantage over NTP, however, the source of this superiority, whether from faster convergence or a superior convergence endpoint remains unclear.
To investigate long-term convergence behavior, we extended training to \textbf{10,000 epochs}.

Drawing from scaling law principles \citep{kaplan2020scalinglawsneurallanguage}, we use small-scale scenarios to infer large-scale behavior. We trained a \textbf{randomly initialized} 438M LLaMA-like model (Ren et al., 2024) on a custom dataset devoid of real-world semantics. This approach eliminates pre-training knowledge effects, allowing pure comparison of NDP and NTP methods. The dataset comprises target items, noise items sharing prefixes with targets, and unrelated items. Noise items simulate real-world interference to target item, while unrelated items help detect overfitting.

Results are presented using similarity between target frequency distributions as a metric, which correlates with loss. 
Figure \ref{fig:similarity_distribution} illustrates that NDP's improved fitting accuracy likely stems from a better convergence endpoint, as NTP fails to close the similarity gap after 10,000 epochs. Both methods achieve over 98\% fitting accuracy on unrelated items, with NDP showing faster initial convergence. These findings suggest that NDP not only converges more rapidly than NTP but also reaches a superior convergence point.

\section{Conclusion}

Our work offers a novel critical perspective on the NTP training paradigm, and this hypothesis was validated through preliminary similarity experiments. Based on addressing this issue, we proposed a new training paradigm inspired from $n$-gram LM called NDP, which achieved good gains in various tasks such as general capability baselines for LLMs, translation, language adaptation, and domain adaptation. Nevertheless, we believe that NDP is merely a simple solution to the narrow candidate problem, and there remains a broad solution space worth exploring to further mitigate this issue.
\bibliography{aaai25}

\section{Reproducibility Checklist}

This paper:
\begin{itemize}
    \item Includes a conceptual outline and/or pseudocode description of AI methods introduced\textbf{ yes}

    \item Clearly delineates statements that are opinions, hypothesis, and speculation from objective facts and results \textbf{yes}

    \item Provides well marked pedagogical references for less-familiare readers to gain background necessary to replicate the paper \textbf{yes}
\end{itemize}

\noindent Does this paper make theoretical contributions? \textbf{yes}

\begin{itemize}
    \item All assumptions and restrictions are stated clearly and formally. \textbf{ yes}

    \item All novel claims are stated formally (e.g., in theorem statements) \textbf{no}

    \item Provides well marked pedagogical references for less-familiare readers to gain background necessary to replicate the paper \textbf{yes}

    \item Proofs of all novel claims are included. \textbf{yes}
    \item Proof sketches or intuitions are given for complex and/or novel results. \textbf{yes}
    \item Appropriate citations to theoretical tools used are given. \textbf{yes}
    All theoretical claims are demonstrated empirically to hold. \textbf{yes}
    All experimental code used to eliminate or disprove claims is included. \textbf{yes}
\end{itemize}

\noindent Does this paper rely on one or more datasets? \textbf{yes}

\begin{itemize}
    \item A motivation is given for why the experiments are conducted on the selected datasets \textbf{yes}
    \item All novel datasets introduced in this paper are included in a data appendix. \textbf{no}
    \item All novel datasets introduced in this paper will be made publicly available upon publication of the paper with a license that allows free usage for research purposes. \textbf{yes}
    \item All datasets drawn from the existing literature (potentially including authors’ own previously published work) are accompanied by appropriate citations. \textbf{yes}
    \item All datasets drawn from the existing literature (potentially including authors’ own previously published work) are publicly available. \textbf{yes}
    \item All datasets that are not publicly available are described in detail, with explanation why publicly available alternatives are not scientifically satisficing. \textbf{yes}
\end{itemize}

Does this paper include computational experiments? \textbf{yes}

\begin{itemize}
    \item Any code required for pre-processing data is included in the appendix.\textbf{ yes}
    \item All source code required for conducting and analyzing the experiments is included in a code appendix. \textbf{yes}
    \item All source code required for conducting and analyzing the experiments will be made publicly available upon publication of the paper with a license that allows free usage for research purposes. \textbf{yes}
    \item All source code implementing new methods have comments detailing the implementation, with references to the paper where each step comes from \textbf{yes}
    \item If an algorithm depends on randomness, then the method used for setting seeds is described in a way sufficient to allow replication of results. \textbf{yes}
    \item This paper specifies the computing infrastructure used for running experiments (hardware and software), including GPU/CPU models; amount of memory; operating system; names and versions of relevant software libraries and frameworks. \textbf{yes}
    \item This paper formally describes evaluation metrics used and explains the motivation for choosing these metrics.\textbf{yes}
    \item This paper states the number of algorithm runs used to compute each reported result. \textbf{yes}

    \item Analysis of experiments goes beyond single-dimensional summaries of performance (e.g., average; median) to include measures of variation, confidence, or other distributional information. \textbf{yes}
    \item The significance of any improvement or decrease in performance is judged using appropriate statistical tests (e.g., Wilcoxon signed-rank). \textbf{yes}
    \item This paper lists all final (hyper-)parameters used for each model/algorithm in the paper’s experiments. \textbf{yes}
    \item This paper states the number and range of values tried per (hyper-) parameter during development of the paper, along with the criterion used for selecting the final parameter setting. \textbf{no}
\end{itemize}

\end{document}

%% file: figures/sharpandsimilarity.tex
\centering
\begin{tikzpicture}[scale=0.45] 

\begin{axis}[
name=plot1, 
at={(0,0)}, 
xshift=-3cm, 
ymajorgrids, 
xmajorgrids, 
grid style=dashed, 
width=.45\textwidth, 
height=.45\textwidth, 
legend style={
at={(1.14,1.04)},
anchor=south,
legend columns=4,
column sep=2.27ex,
font={\Huge},
}, 
xlabel={\Huge Scale(B)}, 
ylabel={\Huge Sharpness}, 
ylabel style={yshift=0em}, 
xlabel style={yshift=0.0em}, 
yticklabel style={/pgf/number format/precision=1, /pgf/number format/fixed zerofill, font={\Huge}}, 
ymin=0.5, ymax=5, 
ytick={1,2,3,4,5}, 
yticklabels={1.0,2.0,3.0,4.0,5.0}, 
yticklabel style={font={\Huge}}, 
xmin=0, xmax=60, 
xtick={0,10,20,30,40,50,60}, 
xticklabels={0,10,20,30,40,50,60}, 
xticklabel style={font={\Huge}} 
]

\addplot[blue!60,mark=pentagon*,mark size=4pt,line width=3pt,mark options={fill=white,draw=blue,line width=1pt}]
coordinates {(0.5,1) (4,1.479851452) (7,2.259327441) (14,3.672399561) (32,4.265417883)};
\addlegendentry{\Huge{Qwen}} 

\addplot[orange!80,mark=triangle*,mark size=5.5pt,line width=3pt,mark options={fill=white,draw=orange,line width=1pt}]
coordinates {(7,1) (12,4.218270426)};
\addlegendentry{\Huge{Mistral}} 

\addplot[teal!70,mark=diamond*,mark size=5pt,line width=3pt,mark options={fill=white,draw=teal,line width=1pt}]
coordinates {(7,1) (13,1.055636607) (70,5.171299311)};
\addlegendentry{\Huge{LLaMA}} 

\addplot[red!70,mark=square*,mark size=4pt,line width=3pt,mark options={fill=white,draw=red,line width=1pt}]
coordinates {(2,1) (7,1.214486961) };
\addlegendentry{\Huge{Gemma}} 

\end{axis}
\node at (plot1.south) [below,yshift=-2.5em] {(a)};

\begin{axis}[
name=plot2,
at={(plot1.east)}, 
anchor=west,
xshift=1.8cm, 
grid style=dashed,
width=.45\textwidth,
height=.45\textwidth,
legend style={at={(1,0.8)}, anchor=south west},
xlabel={\Huge Calibrated Perf.}, 
ylabel={\Huge Similarity Ratio}, 
ylabel style={yshift=-0.8em,xshift=0em},
xlabel style={yshift=-1.4em},
yticklabel style={/pgf/number format/precision=1, /pgf/number format/fixed zerofill,xshift=0.2em, font={\Huge}}, 
ymin=0.998, ymax=1.009, 
ytick={1.00,1.002,1.004,1.006,1.008}, 
yticklabels={1.0,,,,}, 
xmin=0, xmax=12, 
xtick={},
xticklabels={},
legend style={yshift=-50pt, xshift=-50em, legend plot pos=right, font={\Huge}, cells={anchor=west}} 
]
\addplot[gray, line width=2pt, dash pattern=on 10pt off 8pt, domain=0:12] {1};

\addplot[black, line width=3pt, domain=1.4:13, samples=100] {0.9999861351150718 * exp(0.01393066482672535 * x) -0.013941060830111986 * x^1.0143245419780977};

\addplot[only marks, mark=square*, mark size=4pt, color=white, line width=3pt, mark options={fill=none, draw=red!70}]
coordinates {(2.3,1.000107881) (3.9,1.000342486)};

\addplot[only marks, mark=pentagon*, mark size=4pt, color=white, line width=3pt, mark options={fill=none, draw=blue!60}]
coordinates {(9.7,1.004940157) (9.9,1.00530398) (10.6,1.006215302)};

\end{axis}
\node at (plot2.south) [below,yshift=-2.5em] {(b)};

\end{tikzpicture}
\label{fig:data-comparison} 

%% file: figures/domainpicture.tex
\pgfplotsset{
    width=0.8\textwidth, 
    height=0.3\textheight, 
    grid=major, 
    major grid style={dotted}, 
    symbolic x coords={WMT22,IWSLT17,Avg.}, 
    legend style={
        at={(0.5,1.2)}, 
        anchor=north, 
        legend columns=2, 
        column sep=2ex, 
        font=\Huge, 
        inner xsep=4.4ex, 
        cells={anchor=west}, 
    },
}

\begin{tikzpicture}[scale=0.5]
    \begin{axis}[
        ybar, 
        bar width=.8cm, 
        enlarge x limits=0.3, 
        ylabel={\Huge{COMET22}}, 
        xtick=data, 
        ymin=65, ymax=85, 
        ytick={60,65,70,75,80,85}, 
        yticklabels={\Huge{60},\Huge{65},\Huge{70},\Huge{75},\Huge{80},\Huge{85}}, 
        width=0.8\textwidth, 
        height=0.6\textwidth, 
        ylabel style={yshift=0pt,}, 
        xticklabel style={yshift=0pt, font=\Huge}, 
        nodes near coords={\pgfmathprintnumber[fixed,precision=1,zerofill]{\pgfplotspointmeta}}, 
        nodes near coords style={font=\bfseries\large}, 
        every node near coord/.append style={ color=black}, 
    ]
    \addplot+ [fill=white,draw=red, area legend,postaction={pattern=north east lines,pattern color=red}] coordinates {(WMT22,71.3) (IWSLT17,76.4) (Avg.,73.9)};
    \addplot+ [fill=magenta!30,draw=red, area legend,postaction={pattern=crosshatch,pattern color=red}] coordinates {(WMT22,74.2) (IWSLT17,78.9) (Avg.,76.6)};
    \addplot+ [fill=white, draw=blue, area legend,postaction={pattern=north east lines,pattern color=blue!80}] coordinates {(WMT22,77.3) (IWSLT17,80.7) (Avg.,79.0)};
    \addplot+ [fill=cyan!30, draw=blue, area legend,postaction={pattern=crosshatch,pattern color=blue!80}] coordinates {(WMT22,80.3) (IWSLT17,83.0) (Avg.,81.7)};
    \addlegendentry{T5-large(NTP)}
    \addlegendentry{T5-large(NDP)}
    \addlegendentry{T5-xl(NTP)}
    \addlegendentry{T5-xl(NDP)}
    \end{axis}
\end{tikzpicture}